# Fuzzy Clustering Data Given on the Ordinal Scale Based on Membership and Likelihood Functions Sharing

**Zhengbing Hu**
School of Educational Information Technology, Central China Normal University, Wuhan, China
Email: hzb@mail.ccnu.edu.cn

**Yevgeniy V. Bodyanskiy**
Kharkiv National University of Radio Electronics, Kharkiv, Ukraine,
Email: yevgeniy.bodyanskiy@nure.ua

**Oleksii K. Tyshchenko and Viktoriia O. Samitova**
Kharkiv National University of Radio Electronics, Kharkiv, Ukraine,
Email: lehatish@gmail.com, samitova@ gmail.com

*Abstract* — A task of clustering data given on the ordinal scale under conditions of overlapping clusters has been considered. It's proposed to use an approach based on membership and likelihood functions sharing. A number of performed experiments proved effectiveness of the proposed method. The proposed method is characterized by robustness to outliers due to a way of ordering values while constructing membership functions.

*Index Terms*— Computational Intelligence, Machine Learning, ordinal data, FCM, membership function, likelihood function.

## I. Introduction

Data processing tasks that deal with data given not in a numerical form have become really popular nowadays [1, 2]. One can see frequently this sort of tasks in economics, sociology, education and medicine [3-12]. Well-known clustering methods (such as k-means [13, 14], FCM [15, 16], EM-algorithm [17, 18]) usually use an approach based on replacement of linguistic variables by their ranks. But this approach turns out to be incorrect in most cases because it assumes equality of distances between neighboring numerical ranks (which is not always true).

An approach that seems more natural is developed by R.K. Brouwer [19, 20] and based on maximization of a likelihood function. A constraint of this method is an assumption about the Gaussian distribution of initial data which is not fulfilled in many real-world applications as well as a way of likelihood calculation for ordinal variables.

An algorithm of fuzzy clustering data given on the ordinal scale based on membership and likelihood functions sharing is proposed in this article. Initial information for solving this task is an ordered sequence of linguistic variables $x^1, x^2, ..., x^m$, $1 < ... < l-1 < l < l+1 < ...m$ where $x^l$ is a linguistic variable and $l$ is a corresponding rank.

It was introduced in [21-23] to carry out fuzzification procedures for input data based on the occurrence frequency distribution analysis of specific linguistic variables for ordinal data processing. It was also supposed that these distributions were subject to the Gaussian law. An approach was proposed in [24] that was not associated with the hypothesis of normal distribution which will be used in the future work. Thus, initial data for solving the fuzzy clustering task is a sample of images formed by $N$ $n-$dimensional feature vectors $X = \{x(1), x(2), ..., x(k), ..., x(N)\}$ where $X = \{x_1, x_2, ..., x_j, ..., x_N\}$, $j = 1, ..., N$, $x_j = \{x_{jk}^l\}$, $k = 1, ..., n; l = 1, ..., m$ is a rank of a specific value of a linguistic variable in the $k-$th coordinate of the $n-$dimensional space for the $j-$th observation.

A result of this clustering algorithm is partition of an initial data array $X$ into $c$ clusters with membership levels $w_{ij}$ of the $j-$th feature vector to the $i-$th cluster.

The remainder of this paper is organized as follows: Section 2 describes some basic concepts of likelihood and probability. Section 3 describes a fuzzy clustering algorithm based on membership and likelihood functions. Section 4 describes calculation of the conditional probability and the initial data fuzzification. Section 5 presents several synthetic and real-world applications to be solved with the help of the proposed method. Conclusions and future work are given in the final section.

## II. LIKELIHOOD AND PROBABILITY

There are several basic approaches to data clustering such as hierarchical clustering, metric clustering, iterative clustering etc. [25]. Iterative clustering is used in many domains wherein an algorithm detects the best clusters objects may belong to.

Let's consider a rather simple example where each observation has four attributes $x_1, x_2, x_3, x_4$. Let's suppose that they are mutually independent. Then a task of the iteration clustering may be reduced to a task of finding a cluster $y$ via maximization of likelihood $P(y \mid x_1 x_2 x_3 x_4)$ for each observation with features $x_1, x_2, x_3, x_4$. According to the Bayesian formula, this likelihood can be calculated like

$$P(y \mid x_1 x_2 x_3 x_4) = \frac{P(x_1 x_2 x_3 x_4 \mid y)P(y)}{P(x_1 x_2 x_3 x_4)} \quad (1)$$

which means that finding a cluster $y$ via maximization of likelihood $P(y \mid x_1 x_2 x_3 x_4)$ is equivalent to solving this task via maximization of conditional probability $P(x_1 x_2 x_3 x_4 \mid y)$. Moreover, an assumption about the fact that features are mutually independent allows writing down an obvious relation:

$$P(x_1 x_2 x_3 x_4 \mid y) = P(x_1 \mid y)P(x_2 \mid y)P(x_3 \mid y)P(x_4 \mid y). \quad (2)$$

Therefore a problem of finding the cluster $y$ is a problem of maximizing the right side of the equation (2).

Thus, a problem of clusters' finding is solved via maximization of a product of individual conditional probabilities for observation's features. It should be noted that the probability $P(x_j \mid y)$ defines the fact how often an observation $x_j$ appears in a sample with all equal feature values in the cluster $y$. In other words, $P(x_j \mid y)$ is a certain kind of occurrence frequency for $x_j$ with equal parameters' values in the cluster $y$.

## III. A FUZZY CLUSTERING ALGORITHM OF ORDINAL DATA BASED ON MEMBERSHIP AND LIKELIHOOD FUNCTIONS

The proposed algorithm has a rather similar algorithmic structure to FCM.

The clustering task can be solved with the help of FCM for quantitative characteristics by minimization of an objective function

$$Q = \sum_{i=1}^{c} \sum_{j=1}^{N} w_{ij}^{\beta} \left\| x_j - v_i \right\|^2 \quad (3)$$

under constraints

$$w_{ij} \geq 0, \, \forall i = 1,...,c; \, \forall j = 1,...,N;$$
$$\sum_{i=1}^{c} w_{ij} = 1, \, \forall j = 1,...,N; \quad (4)$$
$$\sum_{j=1}^{N} w_{ij} > 0, \, \forall i = 1,...,c$$

where $w_{ij}$ is a membership level of the $j$–th observation to the $i$–th cluster, $\beta$ is a non-negative fuzzifier (a fuzzification parameter). A membership level and clusters' prototypes are calculated according to

$$w_{tj} = \frac{1}{\sum_{i=1}^{c}\left(\frac{\|x_j - v_t\|}{\|x_j - v_i\|}\right)^{\frac{2}{\beta-1}}}, \quad \forall t=1,...,c; \; \forall j=1,...,N \quad \quad (5)$$

$$v_t = \frac{\sum_{j=1}^{N} w_{tj}^{\beta} x_j}{\sum_{j=1}^{N} w_{tj}^{\beta}}, \quad \forall t=1,...,c. \quad \quad (6)$$

One can see from the expressions (5) and (6) that a distance between an observation and a corresponding cluster centroid $v_i$ is used while calculating a membership level $w_{ij}$ of a specific observation to a cluster. Then $v_i$ should be recalculated based on membership levels $w_{ij}$ to clusters. The calculations are carried out iteratively until a stopping condition for the algorithm is met.

An idea of the proposed algorithm consists in the fact that observations' likelihoods are used for defining clusters instead of using distances (like that happens in FCM). Thus, the task is solved via maximization of an objective function

$$Q = \sum_{i=1}^{c}\sum_{j=1}^{N} w_{ij}^{\beta} L_{ij} \quad \quad (7)$$

or minimization

$$Q = \sum_{i=1}^{c}\sum_{j=1}^{N} w_{ij}^{\beta} U_{ij} \quad \quad (8)$$

under constraints

$$w_{ij} \geq 0, \; \forall i=1,...,c; \; \forall j=1,...,N;$$
$$\sum_{i=1}^{c} w_{ij} = 1, \; \forall j=1,...,N; \quad \quad (9)$$
$$\sum_{j=1}^{N} w_{ij} > 0, \; \forall i=1,...,c$$

where $L_{ij}$ stands for a membership likelihood of the $j-$th observation to the $i-$th cluster; $U_{ij}$ stands for a logarithm of dissimilarity for the $j-$th observation with the $i-$th cluster.

The likelihood $L_{ij}$ in (7) is calculated according to

$$L_{ij} = \prod_{k=1}^{n} p_{ijk} \quad \quad (10)$$

where $p_{ijk}$ is a conditional probability for occurrence of a certain value of the $k-$th feature for the $j-$th observation in the $i-$th cluster. It's calculated in the form of

$$p_{ijk} = P(x_{jk} | y_i). \quad \quad (11)$$

The logarithm of dissimilarity in (8) is defined as

$$U_{ij} = -\ln L_{ij} \quad \quad (12)$$

wherein the objective function (8) can be written down in the form

$$Q = \sum_{i=1}^{c}\sum_{j=1}^{N} w_{ij}^{\beta} U_{ij} = \sum_{i=1}^{c}\sum_{j=1}^{N} w_{ij}^{\beta}(-\ln \prod_{k=1}^{n} p_{ijk}) = -\sum_{i=1}^{c}\sum_{j=1}^{N} w_{ij}^{\beta} \sum_{k=1}^{n} \ln p_{ijk}. \qquad (13)$$

To compute $w_{ij}$, one should use an expression [19]

$$w_{tj} = \frac{1}{\sum_{i=1}^{c}\left(\frac{U_{tj}}{U_{ij}}\right)^{\frac{1}{m-1}}}, \quad \forall t = 1,...,c; \; \forall j = 1,...,N. \qquad (14)$$

And membership functions (which are described below in Section 4) should be used for computing conditional probabilities $p_{ijk}$.

A basic drawback of this approach is the fact that an object under consideration is smeared into all existing clusters that leads to a loss of a physical sense in the ordinal scale. In this regard, it seems appropriate to recalculate all distances $d(x_j, v_i)$ after calculating centroids, to put them in an ascending order and to choose the minimal value $d_{\min\min}(x_j, v_i)$ and the next value $d_{\min}(x_j, v_l)$ that follows the minimal one. Considering two minimal distances, we may use these formulas [24]

$$w_{ji} = \frac{d_{\min\min}^{-2}(x_j, v_i)}{d_{\min\min}^{-2}(x_j, v_i) + d_{\min}^{-2}(x_j, v_l)}, \qquad (15)$$

$$w_{jl} = \frac{d_{\min}^{-2}(x_j, v_l)}{d_{\min\min}^{-2}(x_j, v_i) + d_{\min}^{-2}(x_j, v_l)}. \qquad (16)$$

So, the algorithm may be performed as a sequence of steps:

1. Initializing $p_{ijk}$, $\forall i = 1,...,c; \; \forall j = 1,...,N; \; \forall k = 1,...,n$ with random values.
2. Calculating $w_{ij}$, $\forall i = 1,...,c; \; \forall j = 1,...,N$ according to the expression (14).
3. Calculating $p_{ijk}$, $\forall i = 1,...,c; \; \forall j = 1,...,N; \; \forall k = 1,...,n$ with the help of the expression (23).
4. Steps 2 and 3 should be repeated iteratively until a condition is met
$$\varepsilon \leq \max_{ij}\left\{\left|old\_\mu_{ij} - new\_\mu_{ij}\right|\right\}.$$

5. Calculating all the distances $d(x_j, v_i) = \|x_j - v_i\|$ and choosing two minimal distances $d_{\min\min}(x_j, v_i)$ and $d_{\min}(x_j, v_l)$ where $l$ may take on values either $i-1$ or $i+1$;

6. Calculating membership levels for $x_j$ to two neighboring clusters according to the formulas (15) and (16).

IV. CALCULATION OF THE CONDITIONAL PROBABILITY AND THE INITIAL DATA FUZZIFICATION

The fuzzification process for a sequence of rank linguistic variables should be considered on an example of a one-dimensional sample $x_1,...,x_N$ where each observation $x_j$ may be assigned to one of the ranks $l$, $l = 1,...,m$.

Let a value $x_j$ that corresponds to the $l$-th rank occur in the sample $N_l$ times. Then relative occurrence frequencies of the $l$-th rank are taken into consideration

$$f_l = \frac{N_l}{N} \qquad (17)$$

wherein the condition

$$\sum_{i=1}^{m} f_i = 1 \qquad (18)$$

is naturally met.

Averaged occurrence frequencies of observations are formed based on relative frequencies. It's really handy to use a recurrent ratio

$$c_1 = 0.5 f_1,$$
$$c_l = c_{l-1} + 0.5(f_{l-1} + f_l), \forall l = 2,...,m \quad (19)$$

for their calculation.

Then all ordinal data is replaced by the corresponding averaged occurrence frequencies for observations. A fuzzification stage can be presented in the form of a histogram in Fig.1.

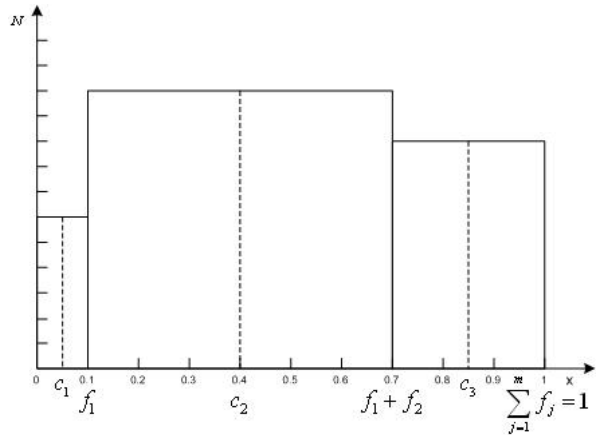

Fig.1. A distribution histogram of ordinal values by occurrence frequency in the sample.

Assuming that a membership level $\mu_{ij} (\forall i = 1,...,c; \forall j = 1,...,N)$ of observations to clusters is known, a mode value is calculated for each feature in each cluster $x_{ik}^*, \forall i = 1,...,c; \forall k = 1,...,n$.

Considering the obtained modes, an asymmetric membership function is built.

1) If $x_{ik}^* > 0.5$, the membership function looks similar to Fig.2 and can be described by a formula

$$\mu_{ijk} = \begin{cases} \dfrac{x_j}{x_{ik}^*}, x \in [0, x_{ik}^*], \\ \dfrac{2x_{ik}^* - x_j}{x_{ik}^*}, x \notin [0, x_{ik}^*]. \end{cases} \quad (20)$$

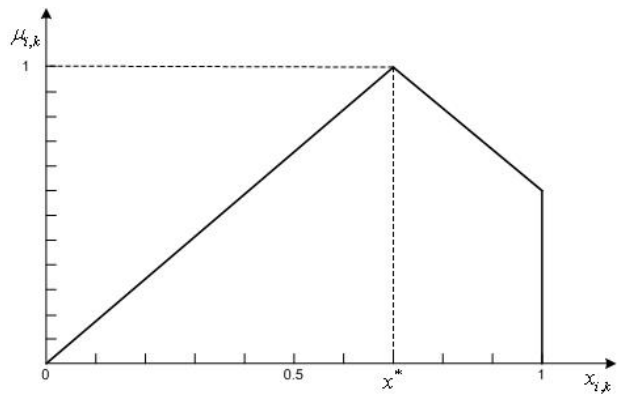

Fig.2. The membership function for a sample of rank values in case of $x_{ik}^* > 0.5$.

2) If $x_{ik}^* < 0.5$, the membership function looks similar to Fig.3 and can be described by a formula

$$\mu_{ijk} = \begin{cases} \dfrac{1-x_j}{1-x_{ik}^*}, x \in [x_{ik}^*, 1], \\ \dfrac{x_j - 2x_{ik}^* + 1}{1 - x_{ik}^*}, x \notin [x_{ik}^*, 1]. \end{cases} \quad (21)$$

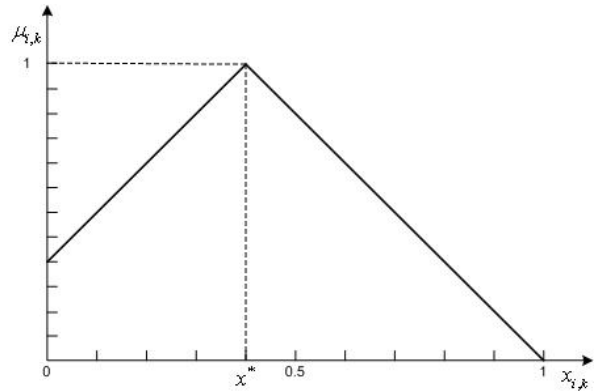

Fig.3. The membership function for a sample of rank values in case of $x_{ik}^* < 0.5$.

3) If $x_{ik}^* = 0.5$, the membership function looks similar to Fig.4 and can be described by a formula

$$\mu_{ijk} = \begin{cases} \dfrac{x_j}{x_{ik}^*}, x \in [0, x_{ik}^*], \\ \dfrac{1-x_j}{1-x_{ik}^*}, x \in [x_{ik}^*, 1]. \end{cases} \quad (22)$$

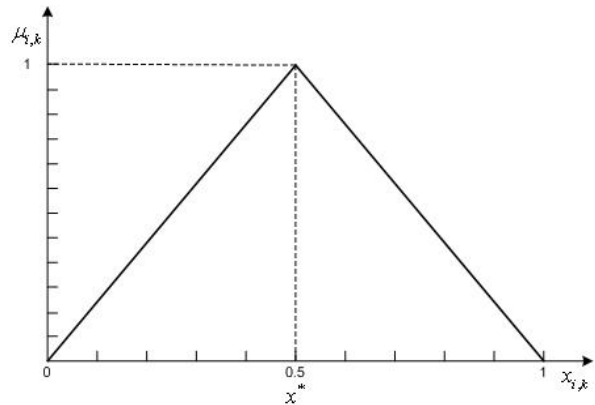

Fig.4. The membership function for a sample of rank values in case of $x_{ik}^* = 0.5$.

Since the conditional probability $p_{ijk}$ directly depends on the occurrence frequency of a specific feature value in the sample and ordinal data is clearly given in a specific order (from the smallest value to the largest one), it may be said that

$$p_{ijk} = \mu_{ijk}. \quad (23)$$

V. EXPERIMENTS

A. Students' performance

To prove the efficiency of the introduced algorithm, we used data describing students' performance. The data contains grades in three courses for 1108 freshmen students in one of Ukrainian universities.

Centroids were determined for each rank (grade) for each value. Results can be found in Table 1.

Table 1. Centroids for observations' ranks

| # of a course/grades | A | B | C | D |
|---|---|---|---|---|
| Course #1 | 0,99 | 0,66 | 0,18 | 0,011 |
| Course #2 | 0,92 | 0,69 | 0,28 | 0,01 |
| Course #3 | 0,96 | 0,75 | 0,29 | 0,006 |

Then the data was divided into four clusters: "Excellent", "Good", "Fair" and "Poor". Clusters' centroids were later obtained as the results of the proposed method (Table 2).

Table 2. Clusters' centroids

| Clusters/parameters | Course #1 | Course #2 | Course #3 |
|---|---|---|---|
| 1 | 0,66 | 0,92 | 0,75 |
| 2 | 0,66 | 0,69 | 0,75 |
| 3 | 0,66 | 0,28 | 0,29 |
| 4 | 0,18 | 0,28 | 0,29 |

15% of students were assigned to the "Excellent" cluster; 26% of students were assigned to the "Good" cluster; 30% of students were assigned to the "Fair" cluster; and 29% of students were assigned to the "Poor" cluster. It would be also interesting to mention that the experiment's results have a 98% match with classification previously performed by the dean's office.

B. Fuzzy clustering ordinal data based on membership and likelihood functions sharing

We chose 3 labeled data samples from the UCI repository for our experiment. A dataset Iris contains 150 observations divided into 3 classes where each observation contains 4 features. A dataset Wine contains 178 observations' vectors divided into 3 clusters where each observation contains 13 features. A dataset Nursery consists of 12958 observations divided into 4 classes where each observation contains 8 features given in the ordinal scale. The datasets Iris and Wine were transformed into the ordinal scale to check the proposed approach. Since every data sample has correct classification labels, clustering efficiency was measured as % of accuracy with respect to benchmark partition after the defuzzification procedure.

For our tests, we selected the FCM algorithm with a parameter $\beta = 2$, a modification of FCM for ordinal data (FCMO) [19] and the proposed fuzzy clustering method (LMFCM). Every cell in Table 3 describes the final result in the form of an average value, a minimum value and a maximum value for a sequence of 50 tests.

The obtained results demonstrate that LMFCM is more stable and its efficiency is higher when compared to FCM and FCMO. Although it should be noted that the results might also depend greatly on a size of a training set.

Table 3. Comparison of clustering accuracy for different data samples

| Methods | Iris | | | Wine | | | Nursary | | |
|---|---|---|---|---|---|---|---|---|---|
| | avg | max | min | avg | max | min | avg | max | min |
| FCM ($\beta =2$) | 70 | 75 | 35 | 69 | 74 | 33 | 62 | 67 | 36 |
| FCMO | 83 | 93 | 58 | 70 | 73 | 45 | 71 | 79 | 43 |
| LMFCM | 85 | 95 | 55 | 71 | 75 | 41 | 74 | 77 | 45 |

C. Automated processing of thermal images for electrical equipment diagnosing

We used the fuzzy clustering method for ordinal data based on membership and likelihood functions sharing to solve a task of automated processing of thermal images for electrical equipment diagnosing. The data was segmented at the first stage and then clustered.

Diagnosis and identification of electrical equipment defects is a topical issue nowadays. Any electrical equipment contains a large number of different elements, contacts and connections. Their weakening, oxidation, load imbalance and overload, burnout wiring insulation, etc. may lead not only to equipment malfunction, but it can also cause an accident.

Timely removal of electrical defects can increase its lifetime and avoid costs for eliminating consequences of an accident.

Thermal imaging diagnostics (a method of non-destructive heat testing) makes it possible to determine the electrical malfunction at early stages of their development. It is based on the analysis of temperature fields which are obtained with the help of portable infrared cameras (thermal imagers). This technique allows carrying out diagnostics while your equipment is on that makes it possible to obtain a more complete picture of existing and emerging defects that are difficult to observe with a naked eye (Fig.5).

Such types of defects can be defined as a violation of electrical elements' encapsulation, overheating of contact connections, deterioration of an internal insulation of windings, a breakdown of elements' sections, inputs' defects, etc. with the help of the thermal imaging diagnosis. Examples of possible defects detected by the thermal imaging diagnostics are shown in Fig.7-10.

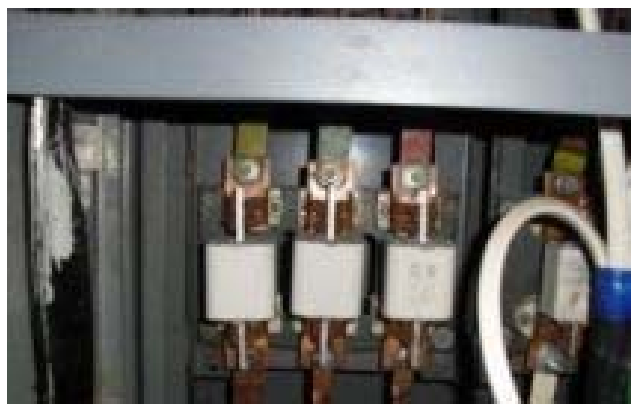

Fig.5. An example of thermal imaging for phase diagnostics in an apparatus of protection and control movement for a mine elevator installation (a photo).

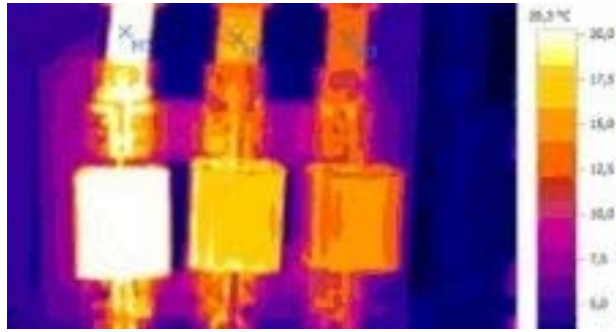

Fig.6. An example of thermal imaging for phase diagnostics in an apparatus of protection and control movement for a mine elevator installation (a thermogram).

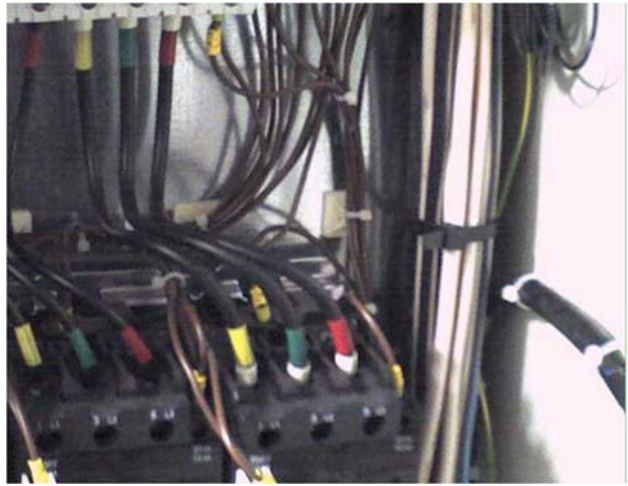

Fig.7. Examples of defects which can be determined during the thermal imaging diagnostics. Exceeding a permissible temperature of a conductor for one phase (a photo).

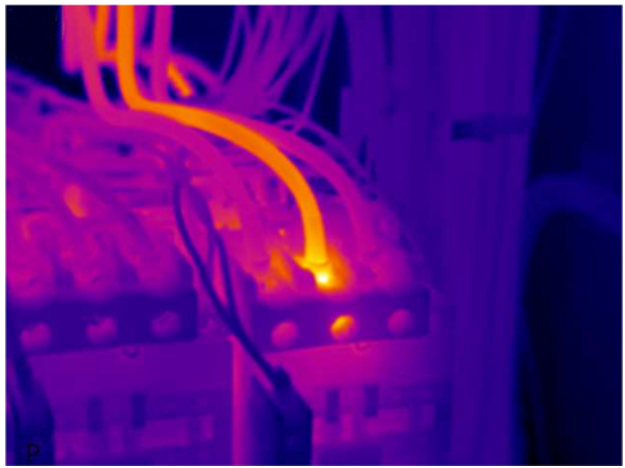

Fig.8. Examples of defects which can be determined during the thermal imaging diagnostics. Exceeding a permissible temperature of a conductor for one phase (a thermogram).

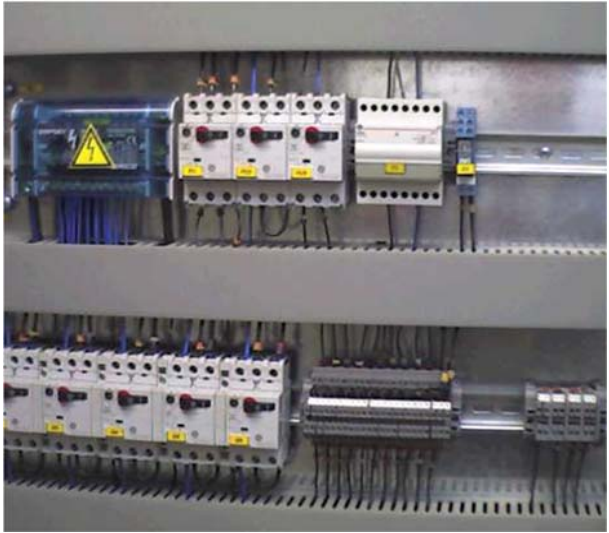

Fig.9. Examples of defects which can be determined during the thermal imaging diagnostics.Overheating of a terminal box (a photo).

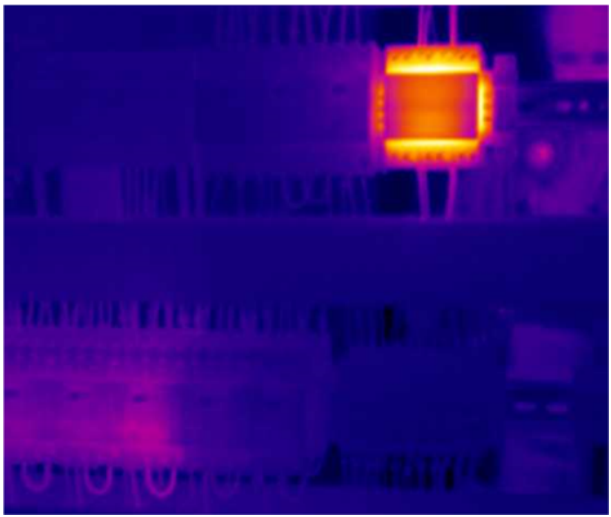

Fig.10. Examples of defects which can be determined during the thermal imaging diagnostics..Overheating of a terminal box (a thermogram).

Some privileges of this approach could be enumerated:

- carrying out diagnostics of electrical equipment in an operating mode (without voltage shutdown);

- detection of defects at early stages of their appearance and development;

- forecasting occurrence of defects;

- low effort required for the diagnostics' production;

- safety of workers during the diagnostics;

- there is no need to arrange a separate workplace;

- an ability to execute a large amount of work in a relatively short period of time;

- high performance and diagnostics' descriptiveness.

Analysis data is a collection of about 5000 thermal images. Some of these thermal images can be seen in Fig.11.

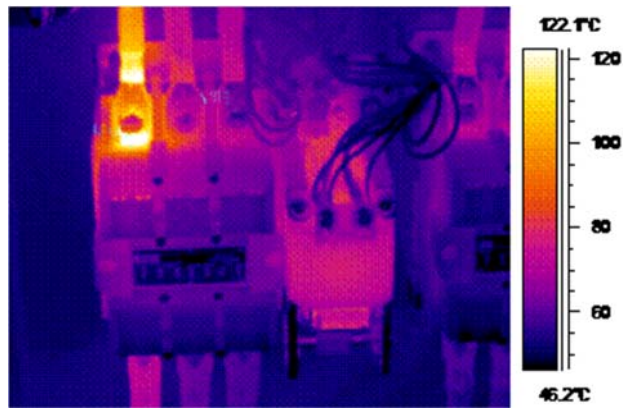

Fig.11. Examples of thermograms.

This problem implies the initial data partition into clusters to determine defects of electrical equipment and to and to develop a sequence of measures for their removal.

The task is solved with the help of two-stage clustering of every image. Images' segmentation is performed at the first stage. An image is represented in a color space $L'a'b'$ for this purpose. Here $L'$ stands for color intensity, $a'$ displays a color of a pixel in the red-and-green axis, and $b'$ displays a pixel's color in the blue-and-yellow axis. After pixelization, an image is a two-dimensional sample which is then divided into clusters using FCM. An example of a thermogram after segmentation is shown in Fig.13.

Ranking is later performed for every pixel of the segmented image according to a thermal image palette of colors that improves accuracy and descriptiveness for further clustering. Moreover, this approach solves a task of a choice of many different palettes during a thermal imaging shooting.

After the ranking procedure for the detected segments is over, the data becomes ordinal. Then the adaptive fuzzy clustering method for ordinal data based on membership and likelihood functions sharing is used.

Implementation of the system made it possible to detect defects of electrical equipment at early stages, increased a speed and efficiency of its diagnostics, and reduced equipment downtime.

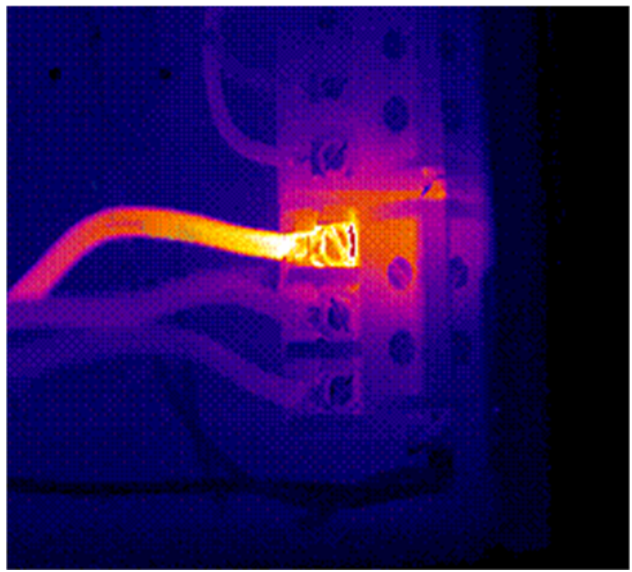

Fig.12. Segmenting a thermogram. An initial view of a thermogram.

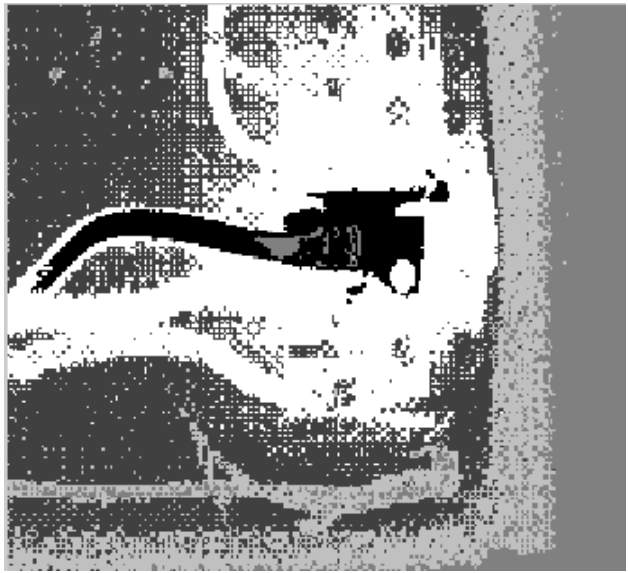

Fig.13. Segmenting a thermogram. A thermogram after segmentation.

VI. Conclusion

The fuzzy clustering algorithm for data given in the ordinal scale based on membership and likelihood functions sharing is proposed. This approach allows effective information processing due to consideration of the nature of the processed data distribution. The fuzzification method and a way of defining a conditional probability $p_{ijk}$ of specific observations' occurrence in every cluster allow quickly and accurately classifying a sample. The main advantage of this method is the fact that it's robust to outliers because of usage of ordered values while constructing membership functions.

References

[1] R. Xu and D.C. Wunsch, *Clustering*. Hoboken, NJ: John Wiley & Sons, Inc. 2009.
[2] C.C. Aggarwal and C.K. Reddy, *Data Clustering. Algorithms and Application*. Boca Raton: CRC Press, 2014.
[3] Zh. Hu, Ye.V. Bodyanskiy, and O.K. Tyshchenko, "A Cascade Deep Neuro-Fuzzy System for High-Dimensional Online Possibilistic Fuzzy Clustering", *Proc. of the XI-th International Scientific and Technical Conference "Computer Science and Information Technologies" (CSIT 2016)*, 2016, Lviv, Ukraine, pp.119-122.
[4] Zh. Hu, Ye.V. Bodyanskiy, and O.K. Tyshchenko, "A Deep Cascade Neuro-Fuzzy System for High-Dimensional Online Fuzzy Clustering", *Proc. of the 2016 IEEE First Int. Conf. on Data Stream Mining & Processing (DSMP)*, 2016, Lviv, Ukraine, pp.318-322.
[5] Ye. Bodyanskiy, O. Tyshchenko, and D. Kopaliani, "An evolving neuro-fuzzy system for online fuzzy clustering", *Proc. Xth Int. Scientific and Technical Conf. "Computer Sciences and Information Technologies (CSIT'2015)"*, 2015, pp.158-161.
[6] Ye. Bodyanskiy, O. Tyshchenko, and D. Kopaliani, "Adaptive learning of an evolving cascade neo-fuzzy system in data stream mining tasks", in *Evolving Systems*, 2016, vol. 7(2), pp.107-116.
[7] Ye. Bodyanskiy, O. Tyshchenko, and D. Kopaliani, "An Extended Neo-Fuzzy Neuron and its Adaptive Learning Algorithm", in *I.J. Intelligent Systems and Applications (IJISA)*, 2015, vol.7(2), pp.21-26.
[8] Ye. Bodyanskiy, O. Tyshchenko, and D. Kopaliani, "A hybrid cascade neural network with an optimized pool in each cascade", in *Soft Computing. A Fusion of Foundations, Methodologies and Applications (Soft Comput)*, 2015, vol. 19(12), pp.3445-3454.
[9] Zh. Hu, Ye.V. Bodyanskiy, O.K. Tyshchenko, and O.O. Boiko, "An Evolving Cascade System Based on a Set of Neo-Fuzzy Nodes", in *International Journal of Intelligent Systems and Applications (IJISA)*, vol. 8(9), 2016, pp.1-7.
[10] Zhengbing Hu, Yevgeniy V. Bodyanskiy, O.K. Tyshchenko, Viktoriia O. Samitova,"Fuzzy Clustering Data Given in the Ordinal Scale", International Journal of Intelligent Systems and Applications(IJISA), Vol.9, No.1, pp.67-74, 2017. DOI: 10.5815/ijisa.2017.01.07.
[11] O. Tyshchenko, "A Reservoir Radial-Basis Function Neural Network in Prediction Tasks", Automatic Control and Computer Sciences, Vol. 50, No. 2, pp. 65-71, 2016.
[12] Ye. Bodyanskiy, O. Tyshchenko, A. Deineko, "An Evolving Radial Basis Neural Network with Adaptive Learning of Its Parameters and Architecture", Automatic Control and Computer Sciences, Vol. 49, No. 5, pp. 255-260, 2015.
[13] Z.B. MacQueen, "Some Methods of Classification and Analysis of Multivariate Observations", *Proc. of the Fifth Berkeley Symposium on Mathematical Statistics and Probability*, 1967, pp.281-297.
[14] Lloyd S. P., Least Squares Quantization in PCM // IEEE Transactions on Information Theory. – 1982. – vol. IT-28. – P. 129-137.
[15] Bezdek J.C., Pattern Recognition with Fuzzy Objective Function Algorithms. – N.Y.:Plenum Press, 1981. – 272p.
[16] Jang J.-Sh. R., Sun Ch.-T., Mizutani E., Neuro-Fuzzy and Soft Computing. – Upper Saddle River, NJ: Prentice Hall, 1997. - 614 p.

[17] Dempster A. P., Laird N. M., and R. D. B., Maximum-Likelihood from Incomplete Data via the EM Algorithm // Journal of the Royal Statistical Society. – 1977. – vol.B. – P. 1-38
[18] Zhong S. and Ghosh J., A Unified Framework for Model-based Clustering // Journal of Machine Learning Research. – 2003. – vol. 4. – P. 1001-1037.
[19] M. Lee and R. K. Brouwer, "Likelihood Based Fuzzy Clustering for Data Sets of Mixed Features", *IEEE Symposium on Foundations of Computational Intelligence (FOCI 2007)*, 2007, pp.544-549.
[20] L. Mahnhoon, "Mapping of Ordinal Feature Values to Numerical Values through Fuzzy Clustering", in *IEEE Trans. on Fuzzy Systems*, 2008, pp.732-737.
[21] Brouwer R.K., Pedrycz W. A feedforward neural network for mapping vectors to fuzzy sets of vectors // Proc.Int.Conf. on Artificial Neural Networks and Neural Information Processing ICANN/ICOMIP 2003. – Istanbul, Turkey, 2003. – P.45-48.
[22] B.S. Butkiewicz, "Robust fuzzy clustering with fuzzy data", in *Lecture Notes in Computer Science*, Vol. 3528, 2005, pp.76-82.
[23] R.K. Brouwer, "Fuzzy set covering of a set of ordinal attributes without parameter sharing", in *Fuzzy Sets and Systems*, 2006, vol. 157(13), pp.1775-1786.
[24] Ye.V. Bodyanskiy, V.A. Opanasenko, and A.N. Slipchenko, "Fuzzy clustering for ordinal data", in *Systemy Obrobky Informacii*, 2007, Iss. 4(62), pp.5-9. (in Russian)
[25] F. Hoeppner, F. Klawonn, R. Kruse, and T. Runkler, *Fuzzy Clustering Analysis: Methods for Classification, Data Analysis and Image Recognition.* Chichester: John Wiley & Sons, 1999.